# From Captions to Rewards (CAREVL): Leveraging Large Language Model Experts for Enhanced Reward Modeling in Large Vision-Language Models

Muzhi Dai[1], Jiashuo Sun[2], Zhiyuan Zhao[1], Shixuan Liu[1],
Rui Li[3], Junyu Gao[1, 4], Xuelong Li[1]

[1]The Institute of Artificial Intelligence (TeleAI), China Telecom, China
[2]Xiamen University, China, [3]Peking University, China
[4]The School of Artificial Intelligence, OPtics and ElectroNics (iOPEN),
Northwestern Polytechnical University, China

## Abstract

Aligning large vision-language models (LVLMs) with human preferences is challenging due to the scarcity of fine-grained, high-quality, and multimodal preference data without human annotations. Existing methods relying on direct distillation often struggle with low-confidence data, leading to suboptimal performance. To address this, we propose CAREVL, a novel method for preference reward modeling by reliably using both high- and low-confidence data. First, a cluster of auxiliary expert models (textual reward models) innovatively leverages image captions as weak supervision signals to filter high-confidence data. The high-confidence data are then used to fine-tune the LVLM. Second, low-confidence data are used to generate diverse preference samples using the fine-tuned LVLM. These samples are then scored and selected to construct reliable chosen-rejected pairs for further training. CAREVL achieves performance improvements over traditional distillation-based methods on VL-RewardBench and MLLM-as-a-Judge benchmark, demonstrating its effectiveness. The code will be released soon.

## 1 Introduction

Large vision-language models (LVLMs) have emerged as critical tools for bridging multimodal understanding, with applications in robotics, content generation, and human-computer interaction (Liu et al., 2023; Bai et al., 2023; Chen et al., 2023; Wang et al., 2024b). A major challenge in deploying these models is aligning their outputs with human preferences, which usually requires post-training alignment. As AI systems increasingly rely on synthetic data for such alignment, accurate evaluations and response rankings become crucial (Liu et al., 2024a; Lambert et al., 2024). Reward modeling addresses this by providing precise preference signals, ensuring mod-els capture human-aligned rewards to boost performance. This reduces dependence on costly human annotations while enabling scalable reinforcement learning and inference-time optimization (Bai et al., 2022; Ouyang et al., 2022). However, robust reward models for vision-language tasks remain underexplored, limiting progress toward human-like multimodal intelligence (Li et al., 2024b).

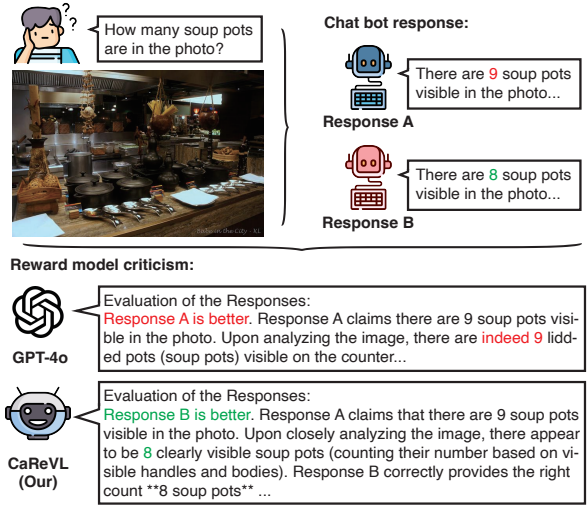

Figure 1: Reward models of LVLM evaluate pairs of responses (A/B) to visual queries and aim to align with human preferences.

LVLM reward models evaluate response pairs (A/B) to visual queries, aiming to align with human preferences closely. Existing methods, like LLaVA-Critic (Xiong et al., 2024) and IXC-2.5-Reward (Zang et al., 2025), rely on oversimplified distillation pipelines that directly replicate preference rankings from stronger models, failing to address synthetic data limitations, such as ambiguous preference boundaries. More specifically, they often either discard low-confidence data entirely or propagate these uncertainties into the reward model, leading to suboptimal performance.

Currently, the primary challenge in LVLM reward modeling remains the scarcity of high-quality and fine-grained preference data for vision-language tasks. An often-overlooked opportunity to address this is leveraging unimodal strengths, particularly the text modality, to enhance error filtering and data curation. Compared to LVLMs, large language models (LLMs) excel at text understanding and preference judgment (Wang et al., 2024a), making them suitable for filtering out inconsistencies in synthetic data. LVLMs can convert visual information into text (e.g., captions), enabling collaboration with LLMs to detect noisy data and retain high-confidence preferences.

We propose CAREVL, a novel method for preference reward modeling that reliably leverages both high- and low-confidence data. First, high-confidence data are collected by filtering GPT-4o (OpenAI et al., 2024a) judgments using auxiliary expert models, all of which are caption-guided textual reward models. The resulting high-confidence data are then used for supervised fine-tuning (SFT). Second, the abandoned low-confidence data are refined and transformed into usable training data. Here, we feed the low-confidence data to the SFT model to sample multiple preference-responses. These samples are then evaluated through a multi-dimensional scoring mechanism, including four aspects: relevance, accuracy, logic, and clarity. Additionally, we introduce a negative sampling strategy, *Best-to-Worse*, to select *chosen-rejected* pairs for Direct Preference Optimization (DPO) (Rafailov et al., 2023). On two benchmarks, VL-RewardBench (Li et al., 2024b) and MLLM-as-a-Judge (Chen et al., 2024a), CAREVL outperforms traditional distillation-based methods, demonstrating its effectiveness.

Our contributions include: (1) a caption-guided data filtering pipeline to produce reliable synthetic preference data; (2) a multi-dimensional scoring mechanism along with the *Best-to-Worse* sampling strategy to enhance preference difference; and (3) a hybrid training strategy allocating high-confidence samples to SFT and ambiguous ones to DPO.

## 2 Related Work

### 2.1 Large Vision-Language Models

LVLMs have progressed significantly in various multimodal tasks, demonstrating strong generalization capabilities (Bai et al., 2023; Wang et al., 2024b; Grattafiori et al., 2024). Recent advances in LVLMs, such as GPT-4V and GPT-4o, have shown remarkable performance in vision-language reasoning, visual dialogue, and multimodal comprehension (OpenAI et al., 2024b,a). An emerging area of research explores using LVLMs as evaluators (Zhang et al., 2023; Liu et al., 2024c). The evaluation can be conducted in both pointwise and pairwise settings, providing a structured framework for preference learning in multimodal applications (Lu et al., 2024b; Yu et al., 2024b).

However, LVLMs still struggle with alignment and preference learning. Many existing models are trained using SFT on extensive datasets but lack an explicit reward mechanism to distinguish between high- and low-quality responses. This limitation underscores the necessity of effective reward modeling for preference alignment in LVLMs.

### 2.2 Reward Modeling

Reward modeling aligns AI models with human preferences, enabling adequate distinction between high- and low-quality responses. Reinforcement Learning from Human Feedback (RLHF) has been used to fine-tune models with explicit preference data (Christiano et al., 2023). However, manual annotations are resource-intensive, and the need for reward models that automatically provide human preference judgments is increasing.

Recent studies have already explored specific LVLM reward models, such as LLaVA-Critic (Xiong et al., 2024) and IXC-2.5-Reward (Zang et al., 2025). Despite these advancements, current methods mainly rely on preference distillation from powerful models, lacking structured techniques to refine and filter reward signals.

## 3 Method

We propose CAREVL to enhance reward modeling in vision-language tasks (Figure 1). CAREVL leverages multiple LVLMs for candidate response generation and uses the combination of a strong LVLM and auxiliary expert models to do pairwise comparisons, obtaining high-confidence data. Then, the high-confidence data are used for SFT, and the low-confidence data are further skillfully designed to be utilized in DPO (Figure 2).

### 3.1 Candidate Response Generation

For a given image-question pair $(I, q)$, we generate candidate responses using $n$ diverse LVLMs (e.g., LLaVA, Qwen2-VL, GPT-4o, etc.). Formally, the set of candidate responses is expressed as:

$$\{r_i\}_{i=1}^n = \text{SampleResponses}(I, q).$$

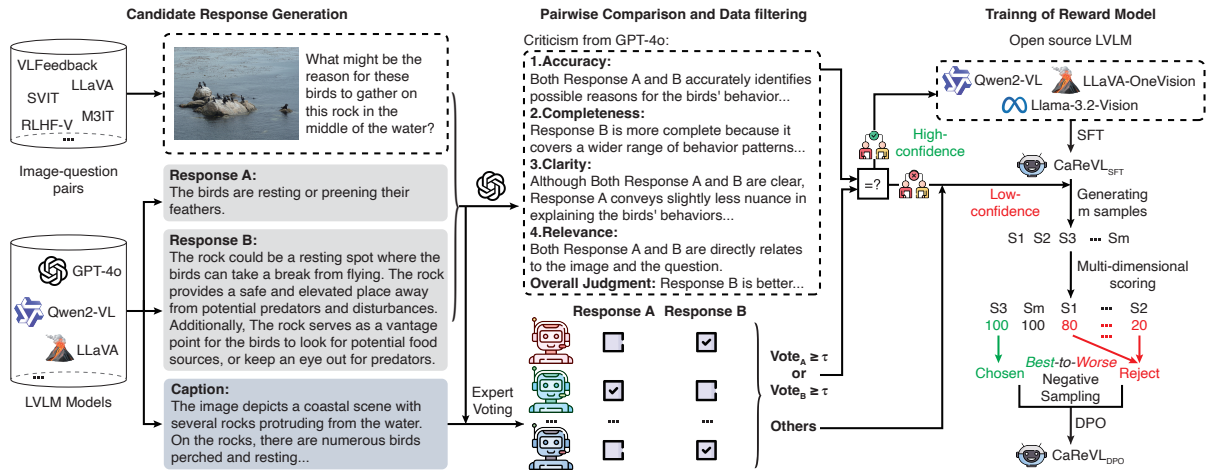

Figure 2: Illustration of CAREVL framework. (1) Candidate Response Generation: CAREVL leverages multiple LVLMs to generate responses (A/B) for each image-question pair; (2) Pairwise Comparison and Data Filtering: GPT-4o and caption-guided auxiliary expert models are used to filter high-confidence data; (3) Training of Reward Model: SFT and DPO are both used to train CAREVL. High-confidence data are used for SFT, and low-confidence data are used for DPO. Before training of DPO, low-confidence data are refined by multiple sampling and multi-dimensional scoring to form chosen-rejected pairs.

This generation is applied to all given $(I, q)$ pairs. The use of multiple models ensures a diverse range of responses in terms of style and content, providing the basis for preference judgments.

### 3.2 Pairwise Comparison and High-Confidence Data Filtering

After generating responses for image-question pairs, we perform pairwise comparisons among them. For two randomly selected $(r_A, r_B)$ with $r_A, r_B \in \{r_i\}_{i=1}^n$ and $r_A \neq r_B$, there are two pairwise comparison processes:

1. A strong LVLM (e.g., GPT-4o) provides an initial binary judgment indicating which response is superior and also provides the explanation.

2. Simultaneously, $M$ auxiliary expert models cast votes on $(r_A, r_B)$. These auxiliary models are purely textual reward models. To endow them with the capability to assess visual tasks, we feed each model the image caption (generated by a powerful LVLM) along with the candidate responses. Denote by $v_A$ the number of votes favoring $r_A$ and by $v_B$ the number of votes favoring $r_B$, such that: $v_A + v_B = M$.

After comparison, we classify data as follows:

- **High-confidence cases:** If one response receives votes meeting or exceeding a threshold, i.e., $v_A \geq \tau$ or $v_B \geq \tau$, and the LVLM's judgment agrees with the majority, the case is labeled as high-confidence data. The threshold is typically set to $M - 1$ ($\tau = M - 1$), ensuring a high level of confidence while tolerating minor disagreements among expert models. These filtered high-confidence data provide strong supervisory signals and are directly utilized for SFT.

- **Low-confidence (ambiguous) cases:** If the vote distribution is balanced (i.e., $\max(v_A, v_B) < \tau$) or conflicts with the LVLM's judgment, the case is considered as low-confidence data. Such low-confidence data need further refinement and are later utilized in DPO.

### 3.3 Supervised Fine-Tuning (SFT)

For high-confidence data, we retain their initial judgment and explanation labels from the strong LVLM. The high-confidence data are used for SFT, enabling LVLM to learn robust and reliable preference patterns for vision-language tasks. The loss function of SFT is as follows.

$$\mathcal{L}_{\text{SFT}} = -\frac{1}{T} \sum_{t=1}^{T} \log \pi(r_t \mid I, q, r_{<t})$$

where $\pi$ is the training model, $(I, q)$ is the image-question pair, $r$ is the response, and $T$ is the total number of tokens in the response. By focusing on high-confidence data, the model avoids overfitting to noisy or ambiguous preference signals

### 3.4 Relabeling for Low-Confidence Data

Ambiguous data often contain diverse and borderline cases that can enrich the preference space. By sampling and relabeling, we refrain from directly using LVLM's low-confidence judgments and explanations, mitigating the risk of propagating noise.

For low-confidence data, we first sample $m$ preference-responses using the SFT model:

$$\{r_j\}_{j=1}^m = \texttt{SamplePreferenceResponses}(I, q)$$

where $m$ is typically set to 10.

Then, preference labels for these samples are re-assigned as follows:

- For data with the vote distribution $\max(v_A, v_B) < \tau$, the label of LVLM's judgment is retained.

- For data with the vote distribution $\max(v_A, v_B) \geq \tau$ but inconsistent with LVLM's judgment, the label is re-assigned based on the majority vote.

If one sample has the same judgment as the re-assigned label, it is regarded as a correct sample. Otherwise, it is an incorrect sample.

### 3.5 *Best-to-Worse* Negative Sampling

To fully use relabeled samples of low-confidence data, we use a multi-dimensional scoring mechanism to automatically evaluate the quality of the samples and also design a *Best-to-Worse* negative sampling strategy, which generates efficient *chosen-rejected* training pairs.

- **Multi-dimensional scoring:** For each sample from the $m$ preference-responses generated by the SFT model, a multi-dimensional scoring mechanism assigns a score $s(r_j)$ to each sample $r_j$. The scoring mechanism is typically an additional expert evaluator (e.g., Llama-3.3-70B with specific prompts) and evaluates samples based on four aspects: relevance, accuracy, logic, and clarity (details in Appendix Table 4 and 5).

- **Determining *chosen* and *rejected* samples:**

  – **Chosen sample ($r^+$):** The highest-scoring correct sample is marked as the chosen sample:

  $$r^+ = r_{\arg\max\{s(r_{\text{correct}})\}_{j=1}^m}$$

  – **Rejected sample ($r^-$):** "All correct samples with scores lower than $r^+$" and "all incorrect samples" are rejected samples:

  $$r^- = \{r_{\text{correct}} \neq r^+\}_{j=1}^m \cup \{r_{\text{incorrect}}\}_{j=1}^m$$

This strategy explicitly emphasizes the distinction between the best sample and all sub-optimal samples, forcing the model to prioritize generating the most preferred outputs during inference.

### 3.6 Direct Preference Optimization (DPO)

The SFT model is finally trained via DPO. Let $\pi(r \mid I, q)$ denote the probability of generating sample $r$ given $(I, q)$ through model $\pi$. We decompose the log-probability components as follows:

$$\mu^+ = \log \frac{\pi_\theta(r^+ \mid I, q)}{\pi_{\text{ref}}(r^+ \mid I, q)}$$

$$\mu^- = \log \frac{\pi_\theta(r^- \mid I, q)}{\pi_{\text{ref}}(r^- \mid I, q)}$$

where $\pi_\theta$ and $\pi_{\text{ref}}$ are both initial from the SFT model, while the training object is $\pi_\theta$ and $\pi_{\text{ref}}$ keeps parameters frozen. In terms of these components, the DPO objective is formulated as:

$$\mathcal{L}_{\text{DPO}} = -\log \sigma\Big(\beta\left(\mu^+ - \mu^-\right)\Big)$$

where $\beta$ is a scaling factor and $\sigma(\cdot)$ denotes the sigmoid function. Minimizing $\mathcal{L}_{\text{DPO}}$ encourages the model to assign higher probabilities to *chosen* samples and decrease probabilities of *rejected* ones.

### 3.7 Inference Phase

Given a new image-question pair $(I, q)$ along with two candidate responses $r_A$ and $r_B$, the generation can be formally represented as:

$$r^* = \texttt{GenerateFinalResponse}(I, q, r_A, r_B),$$

where GenerateFinalResponse denotes the procedure implemented by CAREVL. $r^*$ is the output of the CAREVL, which gives the preference judgment and explanation.

## 4 Experiment

### 4.1 Benchmarks and Evaluation Metrics

We evaluate our method on two LVLM reward modeling benchmarks: VL-RewardBench (Li et al., 2024b) and MLLM-as-a-Judge (Chen et al., 2024a). We follow prior studies and adopt five metrics (*General*, *Hallucination*, *Reasoning*, *Overall Accuracy*, and *Macro Average Accuracy*) for VL-RewardBench and 14 different metrics for MLLM-as-a-Judge (details in Appendix Table 6).

| Models | General | Hallucination | Reasoning | Overall Accuracy | Macro Average Accuracy |
|---|---|---|---|---|---|
| *Proprietary Models* | | | | | |
| GPT-4o-mini (OpenAI et al., 2024a) | 41.7 | 34.5 | 58.2 | 41.5 | 44.8 |
| Claude-3.5-Sonnet (Anthropic, 2023) | 43.4 | 55.0 | 62.3 | 55.3 | 53.6 |
| Gemini-1.5-Flash (Team et al., 2024) | 47.8 | 59.6 | 58.4 | 57.6 | 55.3 |
| GPT-4o (OpenAI et al., 2024a) | 49.1 | 67.6 | **70.5** | 65.8 | 62.4 |
| Gemini-1.5-Pro (Team et al., 2024) | 50.8 | **72.5** | <u>64.2</u> | 67.2 | 62.5 |
| *Open-Source Foundation Models* | | | | | |
| Phi-3.5-Vision (Abdin et al., 2024) | 28.0 | 22.4 | 56.6 | 28.2 | 35.7 |
| InternVL2-8B (Chen et al., 2023) | 35.6 | 41.1 | 59.0 | 44.5 | 45.2 |
| Llama-3.2-90B-Vision (Grattafiori et al., 2024) | 42.6 | 57.3 | 61.7 | 56.2 | 53.9 |
| Molmo-7B (Deitke et al., 2024) | 31.1 | 31.8 | 56.2 | 37.5 | 39.7 |
| Molmo-72B (Deitke et al., 2024) | 33.9 | 42.3 | 54.9 | 44.1 | 43.7 |
| Pixtral-12B (Agrawal et al., 2024) | 35.6 | 25.9 | 59.9 | 35.8 | 40.4 |
| Qwen2-VL-72B (Wang et al., 2024b) | 38.1 | 32.8 | 58.0 | 39.5 | 43.0 |
| NVLM-D-72B (Dai et al., 2024) | 38.9 | 31.6 | 62.0 | 40.1 | 44.1 |
| IXC-2.5-Reward (Zang et al., 2025) | **84.7** | 62.5 | 62.9 | 65.8 | **70.0** |
| LLaVA-OneVision-7B (Li et al., 2024a) | 32.2 | 20.1 | 57.1 | 29.6 | 36.5 |
| LLaVA-Critic-7B (Xiong et al., 2024) | 34.3 | 47.7 | 60.0 | - | 47.4 |
| **CAREVL (LLaVA-OneVision-7B)** | 72.8 | 67.9 | 61.4 | <u>68.7</u> | 67.4 |
| Qwen2-VL-7B (Wang et al., 2024b) | 31.6 | 19.1 | 51.1 | 28.3 | 33.9 |
| **CAREVL (Qwen2-VL-7B)** | 74.2 | 66.3 | 56.3 | 67.8 | 65.6 |
| Llama-3.2-11B-Vision (Grattafiori et al., 2024) | 33.3 | 38.4 | 56.6 | 42.9 | 42.8 |
| **CAREVL (Llama-3.2-11B-Vision)** | <u>76.7</u> | <u>68.8</u> | 60.8 | **70.7** | <u>68.7</u> |

Table 1: The experimental results on VL-RewardBench (Li et al., 2024b) among different backbone models. Most results are directly derived from the original paper. The best results are in **bold**, and the second best results are underlined.

## 4.2 Baselines

We evaluate 25 LVLMs, including both commercial and open-source models. For commercial models, we include prominent options such as GPT-4o/4o-mini (OpenAI et al., 2024a), Claude-3.5-Sonnet (Anthropic, 2023), Gemini-1.5-Flash/Pro (Team et al., 2024), Gemini-Pro (Team et al., 2023), and GPT-4V (OpenAI et al., 2024b). For open-source models, we include Phi-3.5-Vision (Abdin et al., 2024), InternVL2 (Chen et al., 2023), LLaMA-3.2 (Grattafiori et al., 2024), Molmo (Deitke et al., 2024), Pixtral (Agrawal et al., 2024), Qwen2-VL (Wang et al., 2024b), NVLM-D (Dai et al., 2024), IXC-2.5-Reward (Zang et al., 2025), LLaVA-OneVision (Li et al., 2024a), LLaVA-Critic (Xiong et al., 2024), CogVLM (Wang et al., 2023a), and LLaVA-1.5/LLaVA-1.6 (Liu et al., 2023).

## 4.3 Implementation Details

### 4.3.1 Training Data

Our training data comprises about 75k diverse image-question pairs. These pairs are drawn from multiple datasets, including ComVint (Du et al., 2025), LLaVA-Instruction-150k (Liu et al., 2023), LLaVAR (Zhang et al., 2024), LLaVA-Med (Li et al., 2023a), LRV-Instruction (Liu et al., 2024b), M3IT (Li et al., 2023b), SVIT (Zhao et al., 2023), PCA-EVAL (Chen et al., 2024b), VLFeedback (Li et al., 2024c), RLHF (Sun et al., 2023), and RLHF-V (Yu et al., 2024a). We generate the responses of these collected pairs from five ($n$=5) LVLMs, GPT-4o, Qwen2-VL-7B, Qwen2-VL-72B, LLaVA-OneVision-7B, and LLaVA-OneVision-72B, with two sampling strategies (greedy and temperature=1.0). We have checked training data to ensure no overlap with evaluation benchmarks.

We select GPT-4o as the strong LVLM for high-confidence data filtering to generate initial preference judgments and explanations. Furthermore, we integrate five ($M$=5) LLM reward models including INF-ORM-Llama3.1-70B, Skywork-Reward-Gemma-2-27B-v0.2, Skywork-Reward-Gemma-2-27B (Liu et al., 2024a), Skywork-Reward-Llama-3.1-8B-v0.2, and QRM-Llama3.1-8B as auxiliary expert models[1]. To extract caption information, we employ Qwen2-VL-72B. Finally, we use 60K pairs for SFT and 15K for DPO.

### 4.3.2 Training Settings

We choose LLaVA-OneVision-7B, Qwen2-VL-7B, and Llama-3.2-11B-Vision as the backbone models to train CAREVL. Training on three different backbone models allows us to examine the performance of our method more comprehensively. Each of the three backbone models undergoes a two-stage training process: SFT and DPO. During the training, the vision encoders of the LVLMs are frozen.

---
[1] We use expert models from RewardBench (Lambert et al., 2024).

| Settings | Models | COCO | C.C. | Diff. | Graphics | Math | Text | WIT | Chart | VisIT | CC-3M | M2W | SciQA | Aes | MM-Vet | Avg. |
|---|---|---|---|---|---|---|---|---|---|---|---|---|---|---|---|---|
| | | | | | | *Proprietary Models* | | | | | | | | | | |
| | Gemini-Pro (Team et al., 2023) | 0.616 | 0.787 | - | **0.650** | 0.436 | 0.664 | 0.605 | 0.500 | **0.660** | 0.560 | 0.370 | 0.262 | 0.190 | 0.312 | 0.509 |
| | GPT-4V (OpenAI et al., 2024b) | **0.696** | **0.824** | **0.847** | 0.639 | 0.564 | **0.673** | **0.679** | **0.657** | 0.640 | 0.612 | 0.521 | 0.415 | 0.606 | **0.529** | **0.636** |
| | | | | | | *Open-Source Models* | | | | | | | | | | |
| | CogVLM (Wang et al., 2023a) | 0.548 | 0.409 | 0.562 | 0.613 | 0.412 | 0.250 | 0.273 | 0.262 | 0.324 | 0.433 | - | - | - | - | 0.409 |
| | LLaVA-1.5-13b (Liu et al., 2023) | 0.273 | 0.478 | 0.286 | 0.273 | **0.657** | 0.510 | 0.369 | 0.383 | 0.456 | 0.484 | 0.347 | 0.223 | 0.389 | 0.254 | 0.384 |
| | LLaVA-1.6-7b (Liu et al., 2023) | 0.493 | 0.571 | 0.550 | 0.383 | 0.314 | 0.507 | 0.500 | 0.352 | 0.401 | 0.402 | 0.563 | 0.310 | 0.544 | 0.463 | 0.454 |
| | LLaVA-1.6-13b (Liu et al., 2023) | 0.493 | 0.586 | 0.590 | 0.333 | 0.339 | 0.507 | 0.587 | 0.296 | 0.454 | 0.459 | 0.506 | 0.332 | 0.545 | 0.448 | 0.462 |
| | LLaVA-1.6-34b (Liu et al., 2023) | 0.493 | 0.600 | 0.570 | 0.300 | 0.374 | 0.551 | 0.543 | 0.254 | 0.398 | 0.392 | 0.513 | **0.434** | 0.524 | 0.499 | 0.460 |
| Pair w. Tie (↑) | LLaVA-OneVision-7B (Li et al., 2024a) | 0.334 | 0.471 | 0.539 | 0.397 | 0.318 | 0.398 | 0.324 | 0.374 | 0.444 | 0.438 | 0.556 | 0.334 | 0.577 | 0.456 | 0.426 |
| | LLaVA-Critic-7B (Xiong et al., 2024) | 0.593 | 0.687 | 0.707 | 0.587 | 0.432 | 0.544 | 0.564 | 0.338 | 0.596 | **0.628** | 0.591 | 0.370 | **0.686** | 0.464 | 0.556 |
| | CAREVL (LLaVA-OneVision-7B) | 0.569 | 0.673 | 0.760 | 0.608 | 0.465 | 0.544 | **0.607** | 0.326 | 0.588 | 0.608 | **0.645** | 0.322 | **0.695** | 0.388 | 0.549 |
| | Qwen2-VL-7B (Wang et al., 2024b) | 0.384 | 0.417 | 0.247 | 0.422 | 0.331 | 0.356 | 0.254 | 0.310 | 0.398 | 0.471 | 0.346 | 0.416 | 0.332 | 0.393 | 0.365 |
| | CAREVL (Qwen2-VL-7B) | 0.583 | 0.690 | 0.753 | 0.601 | 0.464 | 0.519 | 0.578 | 0.352 | 0.590 | 0.596 | 0.563 | 0.275 | 0.686 | 0.402 | 0.538 |
| | Llama-3.2-11B-Vision (Grattafiori et al., 2024) | 0.298 | 0.303 | **0.890** | 0.207 | 0.286 | 0.411 | 0.404 | 0.468 | 0.370 | 0.386 | 0.539 | 0.308 | 0.542 | 0.506 | 0.393 |
| | CAREVL (Llama-3.2-11B-Vision) | 0.572 | 0.682 | 0.803 | 0.639 | 0.471 | 0.540 | **0.607** | 0.352 | 0.624 | 0.626 | 0.632 | 0.317 | 0.683 | 0.426 | 0.559 |
| | | | | | | *Proprietary Models* | | | | | | | | | | |
| | Gemini-Pro (Team et al., 2023) | 0.717 | 0.840 | - | 0.770 | 0.678 | 0.793 | 0.688 | 0.658 | 0.711 | 0.652 | 0.471 | 0.358 | 0.265 | 0.400 | 0.615 |
| | GPT-4V (OpenAI et al., 2024b) | **0.804** | **0.870** | 0.922 | **0.807** | **0.801** | **0.805** | **0.734** | **0.849** | **0.761** | 0.703 | 0.699 | 0.647 | **0.755** | 0.659 | **0.773** |
| | | | | | | *Open-Source Models* | | | | | | | | | | |
| | CogVLM (Wang et al., 2023a) | 0.654 | 0.450 | 0.643 | 0.704 | 0.481 | 0.292 | 0.500 | 0.423 | 0.500 | 0.591 | - | - | - | - | 0.524 |
| | LLaVA-1.5-13b (Liu et al., 2023) | 0.327 | 0.537 | 0.302 | 0.300 | 0.726 | 0.684 | 0.600 | 0.610 | 0.648 | 0.583 | 0.449 | 0.443 | 0.498 | 0.344 | 0.504 |
| | LLaVA-1.6-7b (Liu et al., 2023) | 0.593 | 0.597 | 0.618 | 0.434 | 0.468 | 0.636 | 0.561 | 0.471 | 0.436 | 0.466 | 0.633 | 0.621 | 0.568 | 0.705 | 0.558 |
| | LLaVA-1.6-13b (Liu et al., 2023) | 0.614 | 0.612 | 0.663 | 0.382 | 0.487 | 0.618 | 0.659 | 0.420 | 0.503 | 0.549 | 0.576 | 0.598 | 0.565 | 0.620 | 0.562 |
| | LLaVA-1.6-34b(Liu et al., 2023) | 0.607 | 0.824 | 0.855 | 0.402 | 0.587 | 0.750 | **0.758** | 0.381 | 0.503 | 0.564 | 0.712 | 0.679 | 0.694 | **0.762** | 0.648 |
| Pair w.o. Tie (↑) | LLaVA-OneVision-7B (Li et al., 2024a) | 0.462 | 0.562 | 0.588 | 0.530 | 0.434 | 0.473 | 0.400 | 0.543 | 0.563 | 0.527 | 0.639 | 0.670 | 0.633 | 0.679 | 0.550 |
| | LLaVA-Critic-7B (Xiong et al., 2024) | 0.771 | 0.774 | 0.755 | 0.758 | 0.596 | 0.658 | 0.600 | 0.488 | 0.727 | **0.742** | 0.692 | 0.658 | 0.715 | 0.635 | 0.689 |
| | CAREVL (LLaVA-OneVision-7B) | 0.744 | 0.751 | 0.797 | 0.776 | 0.624 | 0.637 | 0.728 | 0.472 | 0.711 | 0.712 | **0.725** | 0.652 | 0.716 | 0.577 | 0.690 |
| | Qwen2-VL-7B (Wang et al., 2024b) | 0.358 | 0.389 | 0.224 | 0.441 | 0.290 | 0.346 | 0.208 | 0.220 | 0.367 | 0.472 | 0.309 | 0.305 | 0.317 | 0.300 | 0.329 |
| | CAREVL (Qwen2-VL-7B) | 0.763 | 0.770 | 0.790 | 0.767 | 0.619 | 0.609 | 0.693 | 0.510 | 0.714 | 0.698 | 0.632 | 0.560 | 0.707 | 0.599 | 0.677 |
| | Llama-3.2-11B-Vision (Grattafiori et al., 2024) | 0.390 | 0.338 | **0.934** | 0.264 | 0.383 | 0.482 | 0.485 | 0.678 | 0.447 | 0.453 | 0.605 | 0.628 | 0.558 | 0.753 | 0.494 |
| | CAREVL (Llama-3.2-11B-Vision) | 0.748 | 0.761 | 0.843 | **0.817** | 0.629 | 0.632 | 0.728 | 0.510 | 0.756 | 0.733 | 0.710 | 0.599 | 0.703 | 0.634 | 0.700 |

Table 2: The experimental results on MLLM-as-a-Judge benchmark (Chen et al., 2024a) among different models. Most results are directly derived from the original paper. The best results are in **bold** and the second best results are underlined.

For SFT, we set the learning rate to $1 \times 10^{-5}$, the batch size to 64, and train for 3 epochs. In DPO, we use a learning rate of $1 \times 10^{-6}$, a beta value of 0.01, a batch size of 64, and train for 2 epochs. This ensures a well-tuned reward model.

### 4.4 Main Results

#### 4.4.1 Evaluation on VL-RewardBench

Table 1 summarizes the performance of our approach on the VL-RewardBench, evaluated across five metrics: *General*, *Hallucination*, *Reasoning*, *Overall Accuracy*, and *Macro Average Accuracy*. Our method is applied to three backbone models (LLaVA-OneVision-7B, Qwen2-VL-7B, and Llama-3.2-11B-Vision), and the results demonstrate substantial improvements over all the baseline versions and other competitive models.

For the LLaVA-OneVision backbone, the baseline model (LLaVA-OneVision-7B) achieves an *Overall Accuracy* of 29.6 and a *Macro Average Accuracy* of 36.5. In contrast, our enhanced version of LLaVA-OneVision-7B yields an *Overall Accuracy* of 68.7 and a *Macro Average Accuracy* of 67.4, accompanied by a significant boost in the *General* (72.8 vs. 32.2) and *Hallucination* (67.9 vs. 20.1) metrics. Compared with LLaVA-Critic-7b, which is also trained as a reward model based on LLaVA-OneVision-7B, CAREVL achieves higher performance across all metrics of VL-RewardBench while using less training data (75k vs. 113k), indicating the effectiveness of our method.

Similarly, for Qwen2-VL-7B, CAREVL elevates the *Overall Accuracy* from 28.3 to 67.8 and *Macro Average Accuracy* from 33.9 to 65.6. For Llama-3.2-11B-Vision, CAREVL gets the highest *Overall Accuracy*. Compared to IXC-2.5-Reward, one of the top-performing LVLM reward models on VL-RewardBench, CAREVL exhibits distinct strengths, with both models excelling in different metrics and showcasing complementary advantages. Notably, while IXC-2.5-Reward leverages both open-source and in-house data, CAREVL is trained exclusively on open-source data. Furthermore, CAREVL provides interpretable justifications for preference judgments, whereas IXC-2.5-Reward outputs only numerical scores. These results indicate that our method robustly aligns LVLM outputs with human judgments.

The gain of CAREVL in the *Reasoning* metric is comparatively moderate. This could be the limitation of using "caption-guided auxiliary LLMs" to filter training data. Without direct image input, reasoning scope is confined to a short caption, leading to few high-confidence reasoning data being filtered. Without sufficient high-confidence reasoning data during SFT, it is challenging for CAREVL to learn robust reasoning preference patterns, which may even undermine the effectiveness of subsequent refinement for low-confidence reasoning data. More powerful backbone models or richer high-quality reasoning data may contribute.

|  | General | Hallucination | Reasoning | Overall Accuracy | Macro Average Accuracy |
|---|---|---|---|---|---|
| CAREVL | 74.2 | 66.3 | 56.3 | 67.8 | 65.6 |
| - *Best-to-Worse* Negative Sampling | 54.2 | 55.1 | 56.0 | 54.9 | 55.1 |
| - DPO | 56.1 | 49.8 | 55.0 | 54.3 | 53.7 |
| - Auxiliary Expert Models | 48.7 | 46.5 | 55.3 | 49.8 | 50.1 |

Table 3: The impact of different key components in CAREVL on VL-RewardBench. The backbone model here is Qwen2-VL-7B. The components are progressively removed (indicating as **-**) from top to bottom.

### 4.4.2 Evaluation on MLLM-as-a-Judge Bench

We also present the evaluation of various models on the MLLM-as-a-Judge benchmark. The results, summarized in Table 2, are assessed under two evaluation settings: **Pair w. Tie** and **Pair w.o. Tie**.

CAREVL performs significantly better than the corresponding backbone models in most sub-tasks. Compared to other open-source alternatives, our approach makes progress in the average score (*Avg.*), and shows significant improvements in multiple sub-tasks, such as *Graphics*, *Math*, *WIT*, *M2W*, and *Aes*, highlighting its robustness across different data types.

Since both CAREVL (LLaVA-OneVision-7B) and LLaVA-Critic-7B use the same backbone model, we focus on comparing their performance. Although the two models have their strengths and weaknesses under different data types, CAREVL gets a slightly higher average score (*Avg.*) (0.690 vs. 0.689). Notably, CAREVL uses less training data than LLaVA-Critic (75k vs. 113k), indicating the effectiveness of our method.

Evaluation results indicate a decline in the performance of CAREVL on datasets containing 'tie' type data. This decline is likely due to the absence of such cases in the training data, which biases the model toward making binary preference judgments. In the future, incorporating more 'tie' data could enhance the model performance.

### 4.5 Ablation Studies

#### 4.5.1 Impact of Key Components

To evaluate the contribution of different components in CAREVL, we conduct an ablation study on VL-RewardBench using Qwen2-VL-7B as the backbone model (Table 3). We analyze the effect of cumulatively removing *Best-to-Worse* negative sampling, DPO, and auxiliary expert models on the five metrics. Removing *Best-to-Worse* negative sampling here means that *rejected* samples are randomly selected from incorrect samples.

Removing *Best-to-Worse* negative sampling degrades *General* and *Hallucination* accuracy significantly. Excluding DPO results in a performance decline in all metrics, suggesting that DPO helps generate correct preference judgments. Lastly, removing auxiliary expert models leads to a substantial drop in the *Overall Accuracy* (54.3 to 49.8). These results demonstrate the importance of the components in our method.

#### 4.5.2 Negative Sampling Strategy

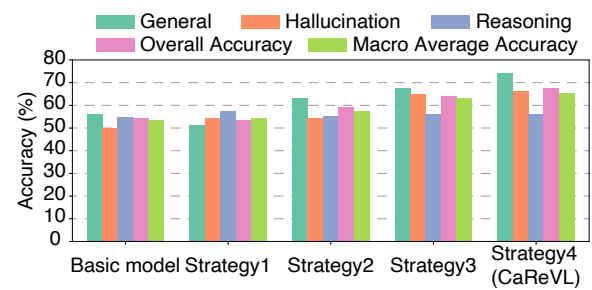

Figure 3: The impact of different negative sampling strategies during DPO on VL-RewardBench. The basic model of DPO is the SFT CAREVL (Qwen2-VL-7B).

Figure 3 presents an ablation study to evaluate the performance of different negative sampling strategies during DPO on VL-RewardBench. We choose one strategy at a time and measure performance across the five metrics. The basic model of DPO is the SFT CAREVL (Qwen2-VL-7B).

Specifically, Strategy1 selects the lowest-scoring incorrect samples as *rejected* samples; Strategy2 chooses the highest-scoring incorrect samples as *rejected* samples; Strategy3 uses all incorrect samples as *rejected* samples; and Strategy4 is our *Best-to-Worse* negative sampling strategy, which uses all the sub-optimal samples (details in Section 3.5). All four strategies use the same *chosen* samples,

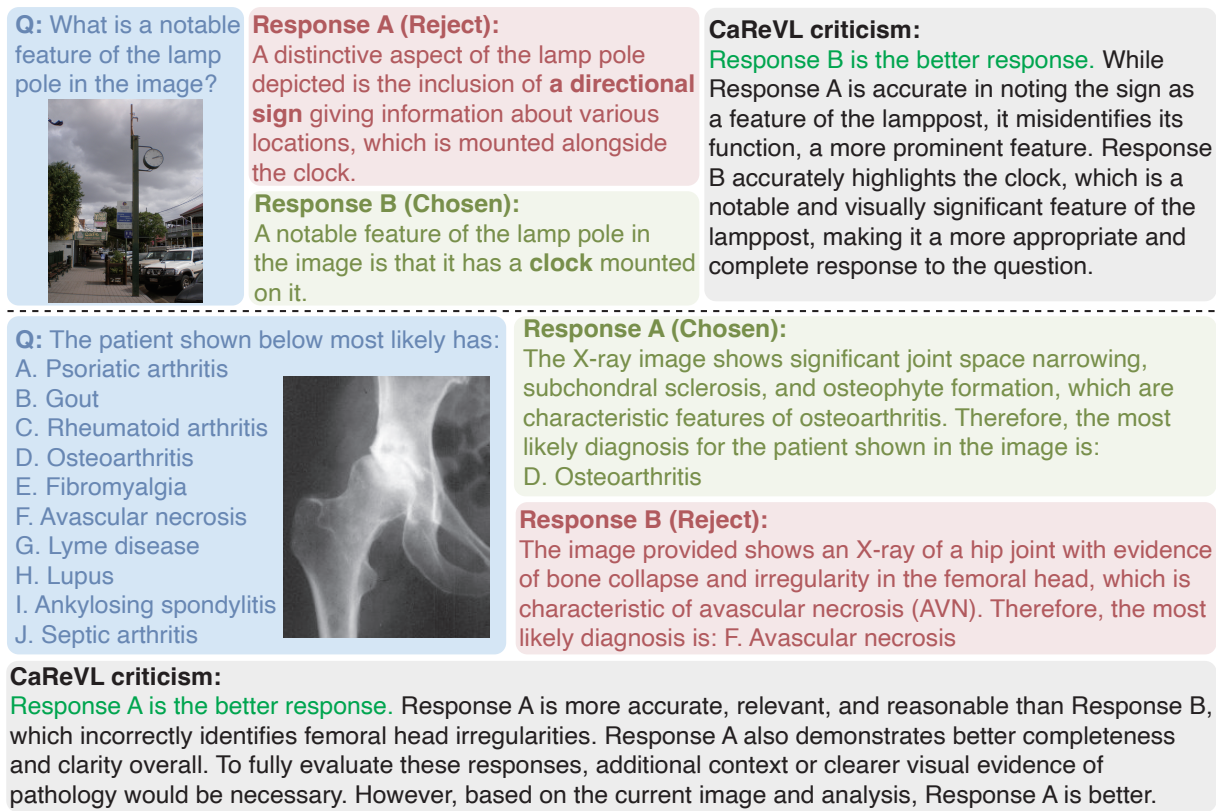

Figure 4: Two examples of CAREVL criticism. The questions, images, and responses are all from VL-RewardBench. The specific process of CAREVL analysis is omitted here, and the core ideas of the criticism are shown.

the highest-scoring correct samples.

Compared with the basic model, Strategy1 has little effect on performance. This may be due to the excessive disparity between the highest-scoring correct samples and the lowest-scoring incorrect samples, which leads to rapid model convergence and hinders CAREVL from fully capturing preference information. In contrast, Strategy2 is better than Strategy1, which also proves that a "high-quality" incorrect sample (hard *rejected* sample) contributes more contrastive information with the *chosen* correct sample than a "low-quality" one.

The marked improvement of Strategy3 and Strategy4 may result from including more prosperous negative signals. Strategy4 takes it further by rejecting samples with low scores, even when they make correct preference judgments. These samples often contain subtle errors, such as minor hallucinations or logical inconsistency, typically overlooked by more straightforward strategies. By rejecting such nuanced cases, Strategy4 includes more challenging negative samples, thereby providing more prosperous negative signals for training. The results in Figure 3 demonstrate the effectiveness of the *Best-to-Worst* negative sampling strategy.

### 4.6 Case Studies

Figure 4 shows two cases of CAREVL from VL-RewardBench. The two cases show that CAREVL can identify errors and select better responses, and the preference outputs are consistent with human intuition. At the same time, CAREVL also provides explanations of preference judgments.

## 5 Conclusion

In this paper, we present CAREVL to align LVLMs with human preferences. Our approach ensures the quality of preference data through (1) leveraging GPT-4o and caption-guided auxiliary expert models to select high-confidence data and (2) a multi-dimensional scoring mechanism and the *Best-to-Worse* negative sampling strategy to refine those low-confidence data. The proposed hybrid training framework effectively utilizes high- and low-confidence data, achieving superior performance compared to traditional distillation-based methods. Experimental results demonstrate the effectiveness of CAREVL, offering new insights into preference alignment for LVLMs.

## 6 Limitations

There are two limitations for CAREVL. As discussed in Section 4.4.1 and Section 4.4.2, our method shows only marginal improvements for reasoning-type data. Additionally, since we do not use "tie" type data during the training process, there is still room for improvement in our method for such response pairs.

## 7 Ethical Considerations

The open-source and commercial models, data, and code used in our work do not involve ethical considerations.

# A Appendix

## A.1 Computation Source

All of our experiments were conducted on eight NVIDIA A800-SXM4-80GB GPUs and a 128-core Intel® Xeon® Platinum 8462Y+ processor. The GPT API model used for data construction is GPT-4o-2024-0513.

## A.2 Statistical Settings

For evaluation, we used the default parameters in the model's built-in generation configuration for inference. In the benchmark evaluation, we performed inference five times for each data, and the final results were averaged to mitigate the impact of randomness.

## A.3 Algorithm

The algorithm for CAREVL is shown in Algorithm 1.

## A.4 Scoring Prompt for CAREVL

The prompts we used in Section 3.5 are shown in Table 4 and Table 5.

## A.5 Dataset composition for MLLM-as-a-Judge benchmark

We show the dataset composition details of the MLLM-as-a-Judge benchmark (Chen et al., 2024a) in Table 6.

**Algorithm 1** Reward Modeling Pipeline for CAREVL
---
**Require:** Dataset $\mathcal{D} = \{(I, q)\}$, number of LVLMs $n$, expert models $M$, threshold $\tau$ (typically $M - 1$), extra samples $m$, DPO iterations $N$
**for** each image-question pair $(I, q) \in \mathcal{D}$ **do**
    Generate candidate responses: $\{r_i\}_{i=1}^n \leftarrow$ SampleResponses$(I, q)$
    **for** each candidate pair $(r_A, r_B)$ from $\{r_i\}$ **do**
        Get strong LVLM judgment $p$ on $(r_A, r_B)$
        **for** $j = 1$ to $M$ **do**
            $Vote_j \leftarrow$ ExpertModel(caption$(I), r_A, r_B)$
        **end for**
        Let $v_A$ be the votes for $r_A$, and $v_B = M - v_A$
        **if** $\max(v_A, v_B) \geq \tau$ **and** $p$ agrees with majority vote **then**
            Mark $(r_A, r_B)$ as **high-confidence**
        **else**
            Mark $(r_A, r_B)$ as **low-confidence**
        **end if**
    **end for**
**end for**
**for** each low-confidence $(I, q)$ **do**
    Generate extra preference responses: $\{r_j\}_{j=1}^m \leftarrow$ SamplePreferenceResponses$(I, q)$
    Relabel based on a majority vote or LVLM judgment:
    **if** votes are ambiguous $(\max(v_A, v_B) < \tau)$ **then**
        Retain LVLM judgment label
    **else**
        Reassign label using majority vote
    **end if**
    Select samples:
$$r^+ = r_{\arg\max\{s(r_{\text{correct}})\}_{j=1}^m}, \quad r^- = \{r_{\text{correct}} \neq r^+\}_{j=1}^m \cup \{r_{\text{incorrect}}\}_{j=1}^m$$

**end for**
**for** $k = 1$ to $N$ **do**
    **for** each pair $(r^+, r^-)$ from low-confidence cases **do**
        Compute $\mu^+ = \log \frac{\pi_\theta(r^+|I,q)}{\pi_{\text{ref}}(r^+|I,q)}$
        Compute $\mu^- = \log \frac{\pi_\theta(r^-|I,q)}{\pi_{\text{ref}}(r^-|I,q)}$
        Update model by minimizing:
$$\mathcal{L}_{\text{DPO}} = -\log \sigma \Big( \beta \left( \mu^+ - \mu^- \right) \Big)$$

    **end for**
**end for**
**Inference:** Given a new image-question pair $(I, q)$ and candidate responses $(r_A, r_B)$, generate the final response:
$$r^* \leftarrow \text{GenerateFinalResponse}(I, q, r_A, r_B)$$

> **Scoring Prompt (Part 1)**
>
> Evaluate the Critique of answers based on the given Question, a corresponding Caption of the image, and two answers to the Question (Answer 1 and Answer 2). Assign a score for the Critic of answers between 0-100 using the following detailed criteria with sub-point scoring guidelines:
>
> 1. Relevance to the Query and Image Content (25 points)
>
> a. (0–5 points): The critique is poorly connected to the query and image content. Either it barely mentions or acknowledges the image and question, or the references are unrelated, generic, or plainly incorrect. The Reference Choice is missing or completely unsubstantiated.
>
> b. (5–10 points): The critique shows minimal relevance to the query and image content. Some connections to the image or question are made but are weak, superficial, or lack depth. The Reference Choice might be present but is vague, incomplete, or not clearly justified.
>
> c. (10–15 points): The critique demonstrates moderate relevance but does not capture all elements of the query and image. While it considers the question and image context, some key points are missed, or the connections are oversimplified. The Reference Choice is included, but explanations are somewhat unclear or lack concrete support.
>
> d. (15–20 points): The critique strongly aligns with the query and image. It discusses both in a reasonably detailed and specific manner, pointing out relevant aspects. The Reference Choice is logical and supported by some evidence, though minor areas for improvement exist in the thoroughness of the explanation.
>
> e. (20–25 points): The critique is perfectly relevant to the query and image content. It effectively bridges the input from the image and question by evaluating Answer 1 and Answer 2. The Reference Choice is highly justified, with detailed reasoning and examples directly tied to image and query elements, showcasing deep understanding.
>
> 2. Factual Accuracy (25 points)
>
> a. (0–5 points): The critique is riddled with factual inaccuracies, completely ignoring critical factual details from the answers or stating incorrect observations. The Critic of Answers fails to identify any errors or inconsistencies between Answer 1 and Answer 2.
>
> b. (5–10 points): The critique contains some factual inaccuracies or overlooks significant factual gaps. Error identification is sporadic or incomplete, and key factual aspects in the answers might be ignored.
>
> c. (10–15 points): The critique includes mostly accurate comments with occasional oversights or minor errors. It identifies and addresses significant factual points and inaccuracies in Answer 1 and Answer 2 but misses some nuanced details.
>
> d. (15–20 points): The critique is factually strong, with excellent consideration of key facts and detailed identification of discrepancies between the two answers. Only minor elements, such as peripheral details or subtle nuances, are missed, but they do not compromise the overall evaluation.
>
> e. (20–25 points): The critique is entirely factually accurate, demonstrating an exhaustive evaluation of the factual correctness of both Answer 1 and Answer 2. It pinpoints inaccuracies, supports insights with evidence, and incorporates nuanced elements without overlooking significant detail.

Table 4: Prompt template for Scoring Part 1.

> **Scoring Prompt (Part 2)**
>
> 3. Logical Reasoning and Analytical Ability (25 points)
> a. (0–5 points): The critique lacks logical reasoning or analytical depth. Conclusions appear arbitrary, unsupported, or even contradictory. There is no clear explanation for why one answer is better than the other.
> b. (5–10 points): The critique shows limited logical reasoning or superficial analysis, offering overly simplistic or poorly constructed arguments. While it attempts to compare Answer 1 and Answer 2, the reasoning is unclear or lacks justification.
> c. (10–15 points): The critique demonstrates moderate reasoning and some degree of sound analysis, but with noticeable gaps. Evaluations of Answer 1 and Answer 2 are generally logical yet may occasionally lack persuasiveness, depth, or detailed support.
> d. (15–20 points): The critique is well-reasoned with structured analytical depth. It evaluates both answers thoroughly and establishes a clear hierarchy between them. Most comparisons are supported with evidence, though minor areas of the reasoning could be refined for greater clarity or rigor.
> e. (20–25 points): The critique is analytically superb, showing exceptional logical reasoning. It precisely deconstructs each element of Answer 1 and Answer 2, clearly explaining the superior answer with detailed, evidence-based arguments. The analysis is nuanced, unbiased, and consistently logical.
> 4. Clarity and Coherence (25 points)
> a. (0–5 points): The critique is poorly written, with significant issues in clarity, structure, and coherence. Ideas are disorganized, hard to follow, or presented ambiguously. Explanations are confusing or largely absent.
> b. (5–10 points): The critique is semi-coherent but lacks fluidity. Its structure may appear disjointed, making it hard to comprehend or follow the critique fully. Explanations are partially clear but include redundancy or confusing phrasing.
> c. (10–15 points): The critique is moderately clear and coherent, though minor difficulty arises in following its flow. It uses a reasonable structure, but transitions or explanations may periodically lack smoothness or full clarity.
> d. (15–20 points): The critique is highly clear and well-structured. It organizes points logically, explaining its reasoning in appropriately detailed and concise terms. Small stylistic or structural improvements could make it truly flawless.
> e. (20–25 points): The critique is perfectly clear, coherent, and concise, with a logical and professional structure. It demonstrates polished writing with excellent transitions, seamless flow, and comprehensive yet straightforward explanations that leave no room for ambiguity. After evaluating the **Critic of answers** based on these criteria, provide the total score in the format: **Score**: <points>
> *Question: { question }*
> *Caption: { caption }*
> *Answer 1: { answer-A }*
> *Answer 2: { answer-B }*
> *Reference Choice: { reference-choice }*
> *Critic of answers: { dpo-sample }*
> *Score:*

Table 5: Prompt template for Scoring Part 2.

| Dataset | Abbreviation | Image Type |
| --- | --- | --- |
| MS COCO (Lin et al., 2015) | COCO | Real-life scene |
| Conceptual Captions (Sharma et al., 2018) | C.C. | Web image |
| DiffusionDB (Wang et al., 2023b) | Diff. | Diffusion |
| InfographicVQA (Mathew et al., 2022) | Graphics | Infographics |
| Math Vista (Lu et al., 2024a) | Math | Mathematics |
| TextVQA (Singh et al., 2019) | Text | Text |
| WIT (Srinivasan et al., 2021) | WIT | Multilingual text |
| ChartQA (Masry et al., 2022) | Chart | Chart |
| VisIT-Bench (Bitton et al., 2023) | VisIT | Comprehensive |
| CC-3M Concept-balanced (Liu et al., 2023) | CC-3M | Comprehensive |
| Mind2Web (Deng et al., 2023) | M2W | Webpage |
| ScienceQA (Lu et al., 2022) | SciQA | Science Knowledge |
| AesBench (Huang et al., 2024) | Aes | Aesthetics Perception |
| MMvet (Yu et al., 2024c) | MM-Vet | Comprehensive |

Table 6: Details of the dataset composition for MLLM-as-a-Judge benchmark (Chen et al., 2024a)